\documentclass[10pt,twocolumn,letterpaper]{article}

\usepackage{cvpr}              

\usepackage{amsmath}
\usepackage{natbib}
\usepackage{enumitem}
\usepackage{graphicx}
\usepackage{algorithm}
\usepackage{algorithmic}
\usepackage{subcaption}
\usepackage[table]{xcolor}
\usepackage{booktabs}
\usepackage{multirow}
\usepackage{array}
\usepackage{newfloat}
\usepackage{listings}

\definecolor{cvprblue}{rgb}{0.21,0.49,0.74}
\usepackage[pagebackref,breaklinks,colorlinks,allcolors=cvprblue]{hyperref}

\DeclareCaptionStyle{ruled}{labelfont=normalfont,labelsep=colon,strut=off}
\floatstyle{ruled}
\newfloat{listing}{tb}{lst}{}
\floatname{listing}{Listing}

\newcolumntype{C}{>{\columncolor{white}}c}
\newcolumntype{G}{>{\columncolor{gray!20}}c}

\def\paperID{*****} 
\def\confName{CVPR}
\def\confYear{2026}

\begin{document}

\title{Preserving Cross-Modal Stability for Visual Unlearning in Multimodal Scenarios}

\author{
    Jinghan Xu \quad Yuyang Zhang \quad Qixuan Cai \quad Jiancheng Chen \quad Keqiu Li\\
    Tianjin University\\
    Tianjin, China
}

\maketitle


\begin{abstract}
Visual modality is the most vulnerable to privacy leakage in real-world multimodal applications like autonomous driving with visual and radar data. Machine unlearning removes specific training data from pre-trained models to address privacy leakage. However, existing methods fail to preserve cross-modal knowledge and maintain intra-class structural stability of retain data, leading to reduced overall and other modalities' performance during visual unlearning. To address these challenges, we propose a \textbf{Cross-modal Contrastive Unlearning (CCU)} framework, which integrates three key components: (a) selective visual unlearning: employing inverse contrastive learning to dissociate visual representations from their original semantics, (b) cross-modal knowledge retention: preserving other modalities' discriminability through semantic consistency, and (c) dual-set contrastive separation: preserving the model performance via isolation of structural perturbations between the unlearn set and retain set. Extensive experiments on three datasets demonstrate the superiority of \textbf{CCU}. Our method achieves a 7.12\% accuracy improvement with only 7\% of the unlearning time compared to the top-accuracy baseline.
\end{abstract}

\section{Introduction}
\label{sec:intro}

\begin{figure}[t]
    \centering
    \subcaptionbox{Performance on test set.\label{subfig:zzt1}}{\includegraphics[width=0.45\columnwidth]{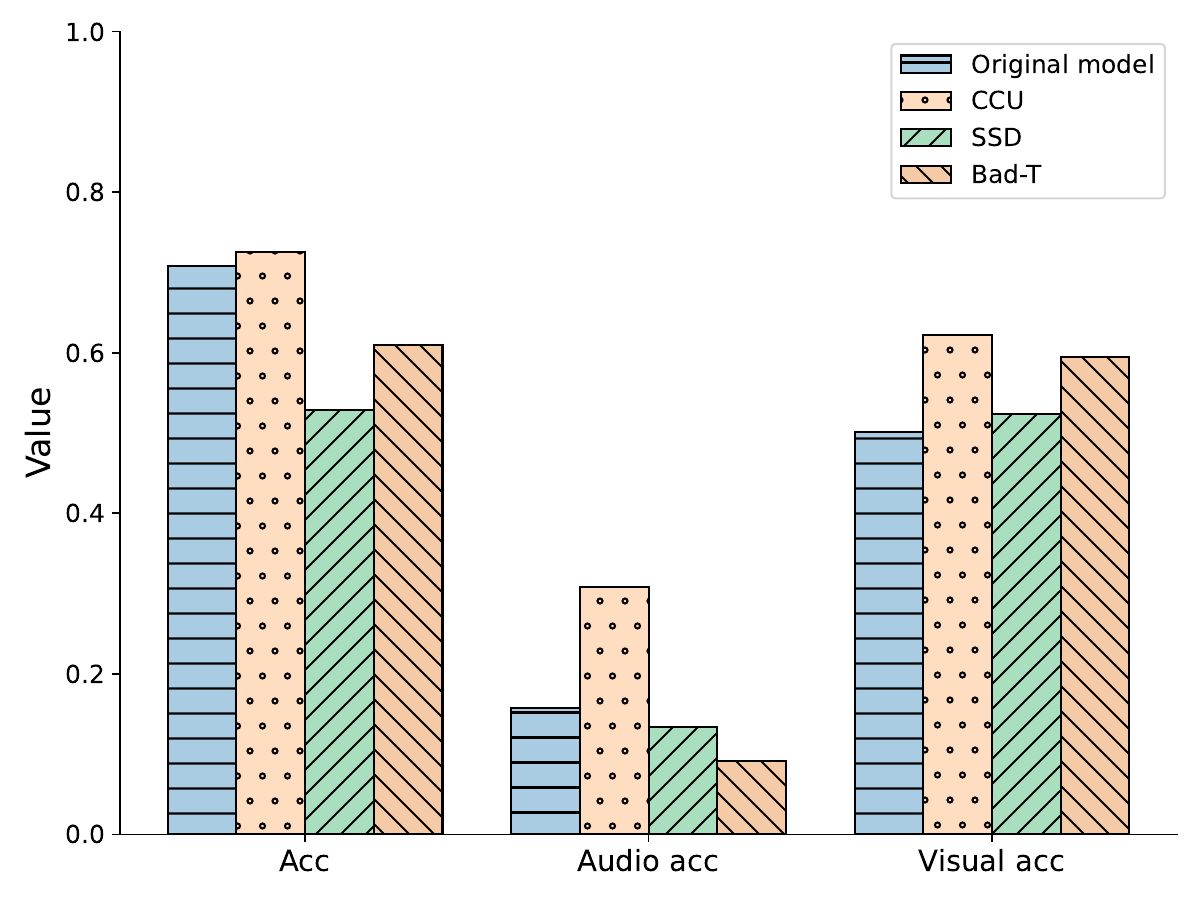}}
    \subcaptionbox{Performance on retain set.\label{subfig:zzt2}}{\includegraphics[width=0.45\columnwidth]{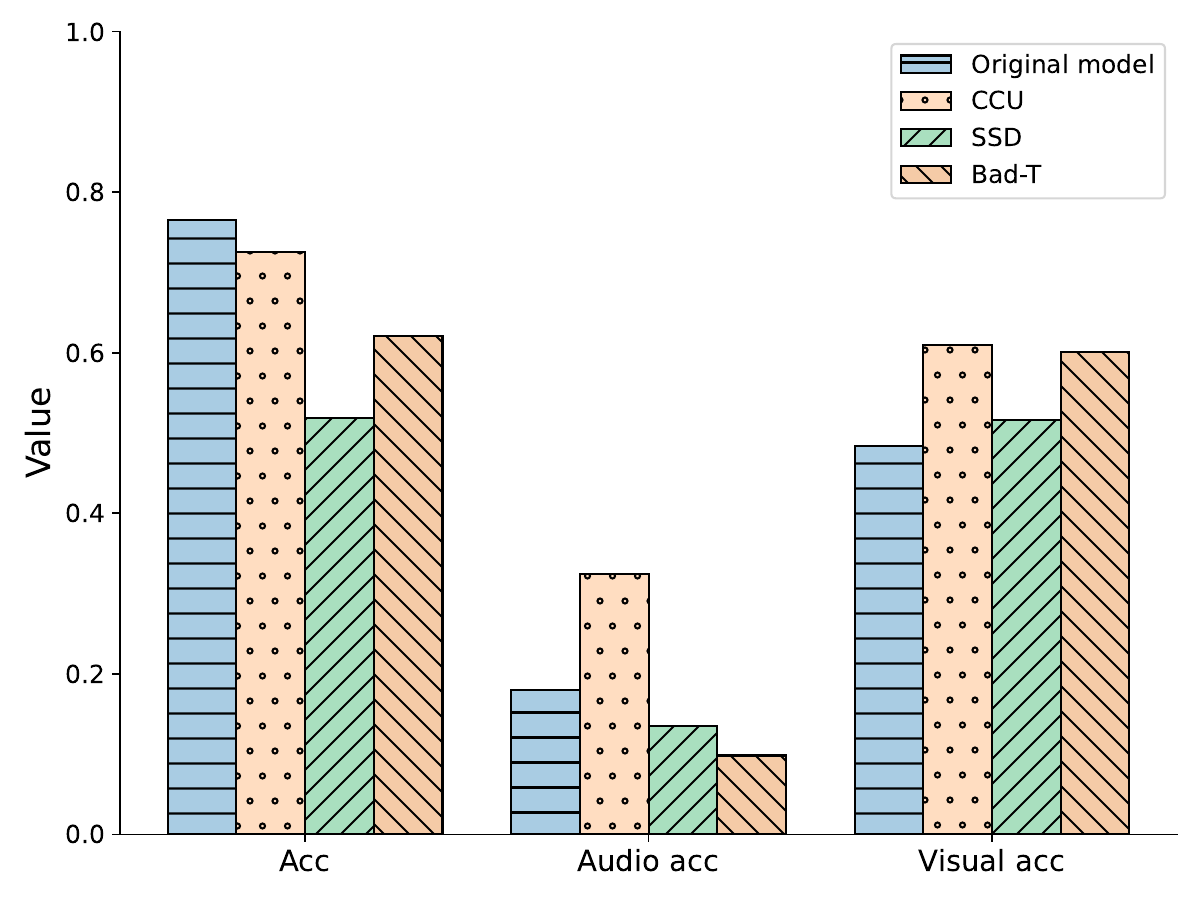}}
    \caption{When using the AV-MNIST \cite{10.1007/978-3-030-11024-6_44} dataset and unlearning 5\% of the visual data in the training set, model performance on the test and retain sets declines significantly, particularly in the audio modality.}
    \label{fig:performance_decline}
\end{figure}

In real-world applications, data often exists in multimodal forms (\textit{e.g.}, image-text, audio-video). For instance, autonomous driving systems process both visual and radar data, while medical diagnosis systems analyze imaging and textual reports. In these scenarios, the visual modality is often the primary source of information but is also the most prone to privacy breaches (\textit{e.g.}, facial recognition, license plate information). Therefore, researching efficient and precise methods to remove the influence of visual data in multimodal scenarios is crucial for complying with privacy regulations like GDPR \cite{voigt2017eu}. Machine unlearning \cite{7163042}\cite{MANTELERO2013229} addresses this challenge by eliminating specific training data's influence from pre-trained models while maintaining overall performance. However, existing unlearning methods, predominantly validated on simple unimodal datasets like MNIST \cite{lecun1998mnist} and CIFAR \cite{krizhevsky2009learning}, they fail to take into account the complex modal dependencies when unlearning the visual modality in multimodal scenarios. As can be seen in Fig. \ref{subfig:zzt1} and Fig. \ref{subfig:zzt2}, unlearning visual modality can degrade the performance of other modalities and the overall performance (\textit{e.g.}, the Bad-T \cite{Chundawat_Tarun_Mandal_Kankanhalli_2023} caused the audio accuracy to drop from about 0.34 to 0.10, which is nearly equivalent to random guessing). While Multidelete \cite{10.1007/978-3-031-72940-9_10} advances the field with its multimodal unlearning framework, it is constrained to concurrent removal of data across all modalities, without the capability for selective unlearning of the visual modality alone. When attempting to unlearn the visual modality in multimodal scenarios, there are mainly the following two challenges: 
\begin{itemize}
    \item Due to the complex interdependencies among modalities, the unlearning of the visual modality will lead to the disruption of cross-modal knowledge, thus affecting the performance of the model in other modalities.
    \item Unlearning the visual modality will cause perturbations to the decision boundary, thereby impacting the overall performance of the model.
\end{itemize}

These methods often fail to take into account the complex modal dependencies when unlearning the visual modality in multimodal scenarios. As can be seen in Fig. \ref{subfig:zzt1} and Fig. \ref{subfig:zzt2}, unlearning visual modality can degrade the performance of other modalities and the overall performance (\textit{e.g.}, the Bad-T \cite{Chundawat_Tarun_Mandal_Kankanhalli_2023} caused the audio accuracy to drop from about 0.34 to 0.10, which is nearly equivalent to random guessing).

To address these challenges, we propose a novel framework named \textbf{Cross-modal Contrastive Unlearning (CCU)}, which can effectively unlearn visual modality while preserving cross-modal knowledge and maintaining model utility. Our \textbf{CCU} framework comprises three dedicated components: (a) the selective visual unlearning module: aiming to decouple visual features of the unlearn set from their original semantic associations, specifically targeting the elimination of privacy-sensitive visual information; (b) cross-modal knowledge retention module: aiming to maintain cross-modal knowledge of the model during visual unlearning, which can effectively counteract the cross-modal interference induced by visual unlearning and preserve the model performance on other modalities; (c) dual-set contrastive separation module: aiming to isolate structural perturbations between the unlearn set and the retain set, ensuring the stability of intra-class representations, which can mitigate the disruption on model performance caused by visual unlearning and preserve the model performance. In summary, our contributions are as follows:

\begin{itemize}
    \item We formalize the visual modality unlearning problem in multimodal scenarios from a representation space perspective and provide theoretical justification for why contrastive learning can effectively eliminate visual semantic associations while preserving essential cross-modal knowledge structures.
    \item We tackle the challenge of maintaining global model performance in complex multimodal systems by explicitly modeling and mitigating bidirectional interactions between the unlearning and retain sets, ensuring robust generalization while preserving critical cross-modal knowledge and eliminating the influence of unlearned visual data.
    \item We conduct extensive experimental validation on multiple datasets, demonstrating the superiority of our approach in balancing unlearning effectiveness, model utility, and computational efficiency.
\end{itemize}

In Section 2, we present some related work of machine unlearning. In Section 3, we describe the problem definition of visual unlearning in multimodal scenarios. In Section 4, we describe the details of our framework. Finally, Section 6 concludes this paper.

\section{Related Work}
\subsection{Exact Machine Unlearning}
The fundamental approach for exact machine unlearning involves removing the unlearn set and retraining the model from scratch, though this incurs substantial computational costs, particularly for deep learning models and complex datasets \cite{9797378}. Consequently, researchers have developed more efficient strategies.

Research \cite{7163042} reformulated traditional ML algorithms into summation forms, requiring only incremental updates to summations during unlearning operations to achieve faster execution than full retraining. SISA \cite{9519428} partitions training datasets into shards and trains separate models on each subset, enabling localized retraining of affected shards during unlearning. PUMA \cite{Wu_Hashemi_Srinivasa_2022} establishes a framework for precisely modeling individual training samples' impacts across performance metrics and systematically eliminating targeted samples' influence. Golatkar et al. \cite{9577384} developed a deep network unlearning method by decomposing models into data-independent core components and learnable parameter-bound sections. While computationally efficient, these methods demand significant storage resources for intermediate training parameters and partitioned subsets, with model performance being heavily dependent on partitioning strategies \cite{10136160}.

\subsection{Approximate Machine Unlearning}
Approximate unlearning produces models that closely mimic those trained on residual data, eliminating the need for retraining. Unlike exact unlearning, which ensures model equivalence, approximate methods estimate and mitigate the influence of unlearned samples. Key approaches include:

Finetune \cite{warnecke2021machine} leverages catastrophic forgetting through retain set fine-tuning. NegGrad \cite{9157084} implements unlearning via loss maximization on target samples, though with notable performance degradation. Bad-T \cite{Chundawat_Tarun_Mandal_Kankanhalli_2023} employs a dual-teacher framework where an incompetent teacher propagates misleading information about unlearned data while a competent teacher preserves valid knowledge. SCRUB \cite{10.5555/3666122.3666217} formulates unlearning as behavioral mimicry optimization, enforcing compliance on retained data and divergence on unlearned samples. SSD \cite{10.1609/aaai.v38i11.29092} utilizes Fisher Information Matrix analysis to identify and selectively dampen parameters critical to unlearned samples. Zhang et al. \cite{zhang2024contrastive} propose contrastive unlearning by optimizing embedding space similarities between unlearn samples and class-defined positive/negative pairs, though bidirectional interference between unlearning and retain sets retains unaddressed. Current limitations include predominant validation on unimodal datasets (MNIST/CIFAR) and limited multimodal adaptability. Multidelete \cite{10.1007/978-3-031-72940-9_10} introduces multimodal unlearning through modality decoupling and knowledge retention losses, yet it is only applicable for removing data across all modalities and cannot be used to selectively unlearn the visual modality alone.

\section{Problem Definition}
In this section, we aim to make the first step towards the definition of visual unlearning in multimodal scenarios. Consider a multimodal model $f: V \times A \to Y$ trained on dataset $D_{\text{train}} = \{(v_i, a_i, y_i)\}_{i=1}^N$, where $v_i \in V$ (visual inputs, e.g., images), $a_i \in A$ (audio inputs, e.g., waveforms), and $y_i \in \mathbb{R}^K$ (semantic labels, e.g., class probabilities). Given a unlearning subset $D_f \subset D_{\text{train}}$ and its complement $D_r = D_{\text{train}} \setminus D_f$, we decompose the model into three components:
\begin{equation}
    f = (f_V, f_A, f_F) \quad \text{with} \quad 
    \begin{cases}
        f_V: V \to \mathbb{R}^d & \\
        f_A: A \to \mathbb{R}^d & \\
        f_F: \mathbb{R}^d \times \mathbb{R}^d \to Y & 
    \end{cases}
\end{equation}
where $f_V$ denotes a visual encoder, $f_A$ denotes an audio encoder, and $f_F$ denotes a fusion encoder. Our target is to construct a desired unlearned model $f' = (f'_V, f'_A, f'_F)$, which behaves as if the visual data subset $V_f$ had never been used during the training of the original model $f$. The goal is to make the samples in \(V_f\) closer to the samples of other classes in \(V_r\) and farther from the samples of its own class in \(V_r\) in the representation space, thus approaching the decision boundary. At the same time, to ensure the performance of the model, the data not involved in the unlearning process should maintain its original representation as much as possible, minimizing the perturbation of the decision boundary caused by unlearning. By doing so, the model can effectively ``unlearn'' the patterns learned from $V_f$. Crucially, we strive to ensure that the performance of the unlearned model $f'$ on the test data $D_{test}$ is as close as possible to that of the original model $f$. Moreover, the impact on other modalities should be minimized. This approach will guarantee that while $f'$ discards specific data knowledge from $V_f$, it preserves the overall task effectiveness of $f$ on the retain set $D_r$. Ideally, $f'$ should be able to effectively unlearn $V_f$ without degrading its performance on $D_{test}$ and $D_{r}$, and without significantly affecting performance on other modalities.. 
\begin{equation}
    \text{Acc}(f',D_{test}) \approx \text{Acc}(f,D_{test}) \approx \text{Acc}(f',D_{f})
\end{equation}
\begin{equation}
    \text{Acc}(f',D_{r}) \approx \text{Acc}(f,D_{r})
\end{equation}
Notably, the unlearning process is strictly confined to visual components, leaving other modalities unaffected.

\section{Methodology}
To address visual unlearning in multimodal scenarios, we propose a versatile framework for selective visual unlearning, implemented via vision-audio modality pairing while retaining extensibility to other pairings, such as vision-text. Figure~\ref{fig1} illustrates our framework, with the algorithm's pseudocode provided in the appendix.

\begin{figure*}[t]
\centering
\includegraphics[width=0.95\textwidth]{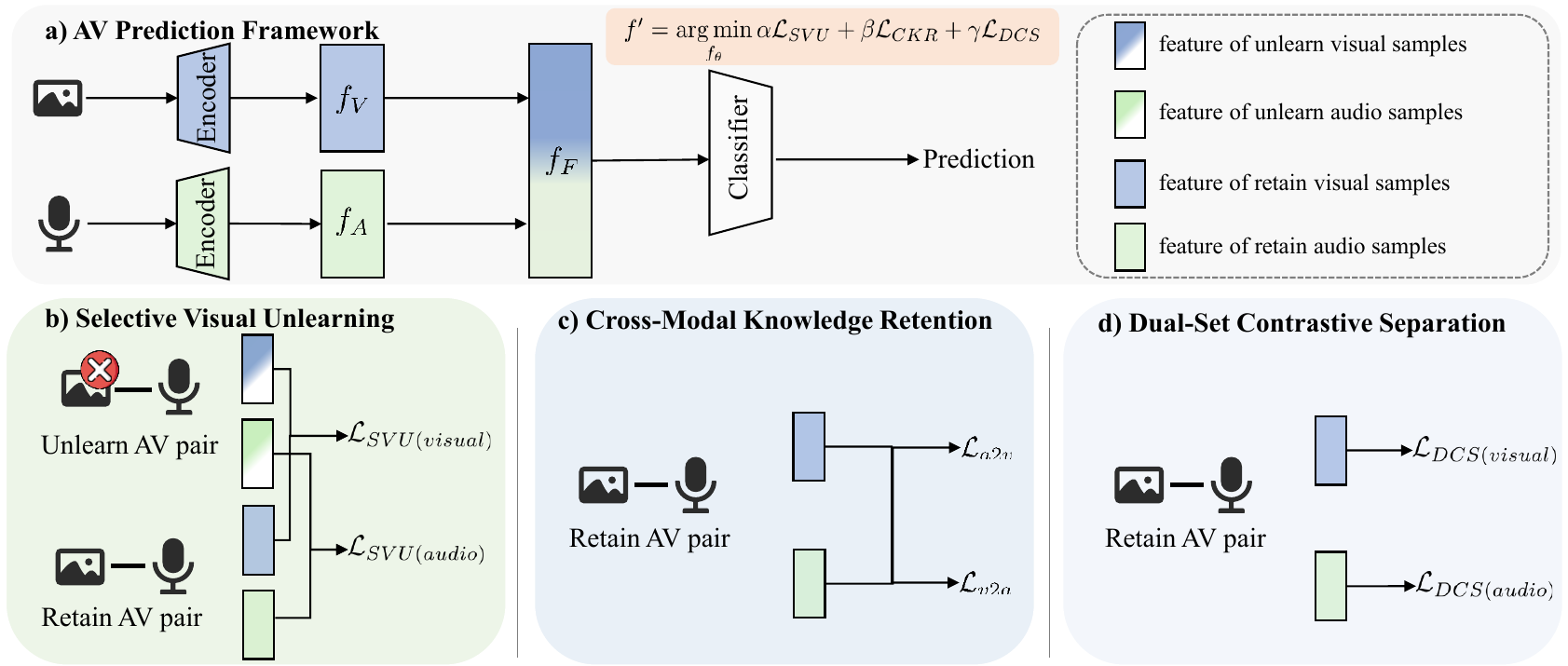}
\caption{CCU framework. It contains three modules: (b) selective visual unlearning, (c) cross-modal knowledge retention and (d) dual-set contrastive separation. We take visual-audio learning for example to illustrate our framework and we only unlearn visual samples. Selective visual unlearning module uses features of AV pair from unlearn set and retain set to reach the goal of visual unlearning. Cross-modal knowledge retention module takes features of AV pair from retain set to mitigate the cross-modal knowledge disruption caused by selective visual unlearning. Dual-set contrastive separation module utilizes features of AV pair from retain set to preserve the stability of intra-class representations and ensure robust model performance. The combination can achieve our goal of visual unlearning while preserving model performance.}
\label{fig1}
\end{figure*}

\subsection{Selective Visual Unlearning}
The selective visual unlearning module eliminates visual modality knowledge while preserving audio performance. Each training iteration processes an unlearning batch ${X}_u = \{(v_i, a_i, y_i)\}_{i=1}^B$ with size $B$, which is sampled from the unlearn data $D_f$, and a retain batch ${X}_r = \{(v_j, a_j, y_j)\}_{j=1}^B$ is sampled from the retain set $D_r$. We denote $v_i$ and $a_i$ in $\{v_i, a_i\}$ as anchors. Based on those anchors, positives and negatives are chosen from $X_r$. For visual modality, positives are ${P}(v_i) = \{v_j \mid (v_j, a_j, y_j) \in {X}_r, y_j = y_i\}$; negatives are ${N}(v_i) = \{v_j \mid (v_j, a_j, y_j) \in {X}_r, y_j \neq y_i\}$. For audio modality, the construction of positives ${P}(a_i)$ and negatives ${N}(a_i)$ is similar. Correspondingly, let embeddings of positives and negatives of visual modality be $P_{m}(v_i) = \{m_j \mid m_j = f_V(v_j), v_j \in P_v(v_i)\}$ and $N_{m}(v_i) = \{m_j \mid m_j = f_V(v_j), v_j \in N_v(v_i)\}$, and the positives $P_{n}(a_i)$ and negatives $N_{n}(a_i)$ of audio modality are similar. The purpose of the SVU loss function in Eq. (\ref{con:SVUloss}) is to minimize the similarity of positive sample pairs and maximize the similarity of negative sample pairs in the visual modality, while maximizing the similarity of positive sample pairs and minimizing the similarity of negative sample pairs in the audio modality.
\begin{equation}
\resizebox{1\columnwidth}{!}{$%
\begin{split}
    \mathcal{L}_{SVU} = & - \sum_{v_i \in {V}_u} \frac{1}{|N_m(v_i)|} \sum_{m_a \in N_m(v_i)} \log \frac{\exp(m_i \cdot m_a / \tau_{S})}{\sum_{m_p \in P_m(v_i)} \exp(m_i \cdot m_p / \tau_{S})} \\
    & - \sum_{a_i \in {A}_u} \frac{1}{|P_n(a_i)|} \sum_{n_p \in P_n(a_i)} \log \frac{\exp(n_i \cdot n_p / \tau_{S})}{\sum_{n_a \in N_n(a_i)} \exp(n_i \cdot n_a / \tau_{S})}
\end{split}
$}
\label{con:SVUloss}
\end{equation}

$\tau_{SVU} \in \mathbb{R}^+$, which is in Eq. (\ref{con:SVUloss}), is a scalar temperature parameter. In our final algorithm, we contrast each $X^u$ with $\omega$ randomly sampled batches $X_r$. Thus, within a single unlearning round, our algorithm processes each batch of data from $D_{r}$ for $\omega$ times.

\subsection{Cross-Modal Knowledge Retention}
To mitigate cross-modal knowledge degradation during visual unlearning, we design a knowledge retention loss which draws inspiration from CLIP's contrastive paradigm \cite{pmlr-v139-radford21a}. This module explicitly aligns visual and audio modalities by leveraging their latent semantic correlations. Specifically, for samples in the retain set, we compute feature embeddings for paired visual and audio data, and introduce a cross-modal contrastive loss to optimize their joint representation space. $V_{m} = [m_1,...,m_B] $ denotes the embeddings of visual data in the retain set. $A_{n} = [n_1,...,n_B]$ denotes the embeddings of audio data in the retain set. $S_{i,j} = \frac{A_{n}^TV_m}{\tau_{CKR}}$ denotes the similarity matrix computed by $A_{n}$ and $V_{m}$. Then we define cross-modal knowledge retention loss functions as follows:
\begin{equation}
    \mathcal{L}_{\text{CKR}} =   -\underbrace{\sum_{i=1}^{B} \log \frac{e^{S_{i,i}}}{\sum_{j = 1}^{{B}}e^{{S}_{i,j}}}}_{\mathcal{L}_{a2v}} - \underbrace{\sum_{j=1}^{B} \log \frac{e^{S_{j,j}}}{\sum_{i = 1}^{{B}}e^{{S}_{j,i}}}}_{\mathcal{L}_{v2a}}
\end{equation}

This dual optimization simultaneously maximizes alignment for matched pairs while discouraging mismatched associations, maintaining discriminative cross-modal representations throughout the unlearning process. The formulation ensures semantic consistency between modalities by reinforcing shared knowledge structures in the embedding space.

\subsection{Dual-Set Contrastive Separation}

\begin{figure}[ht!]
    \centering
    \begin{subfigure}[t]{0.49\linewidth}
        \centering
        \includegraphics[width=0.9\linewidth]{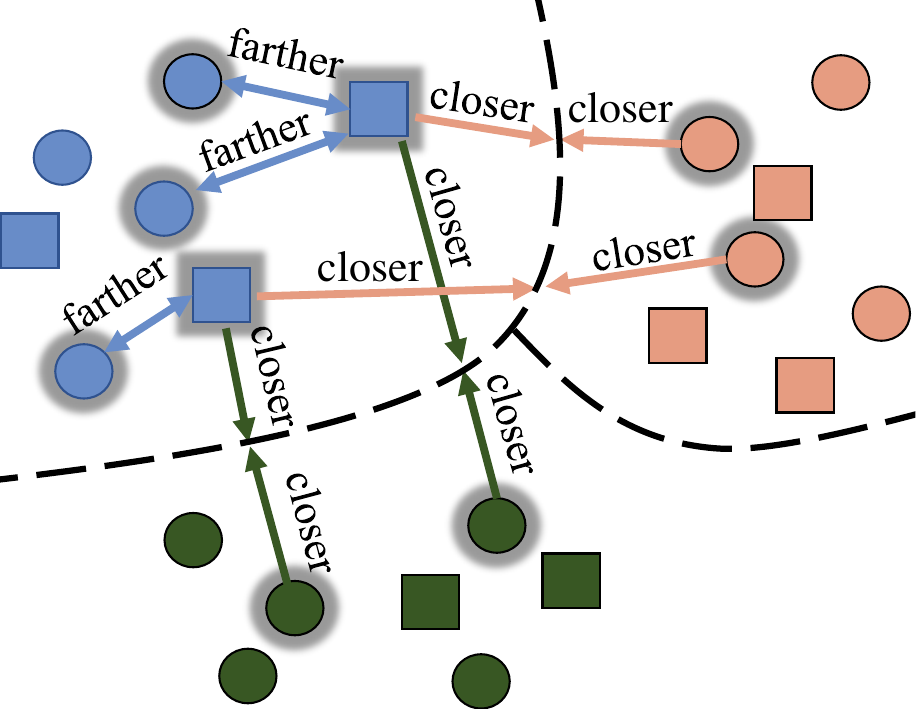}
        \caption{Visual modality representation spaces prior to selective visual unlearning, which increases similarity between unlearn samples and different classes in the current batch.}
        \label{chutian1}
    \end{subfigure}%
    \hfill
    \begin{subfigure}[t]{0.49\linewidth}
        \centering
        \includegraphics[width=0.9\linewidth]{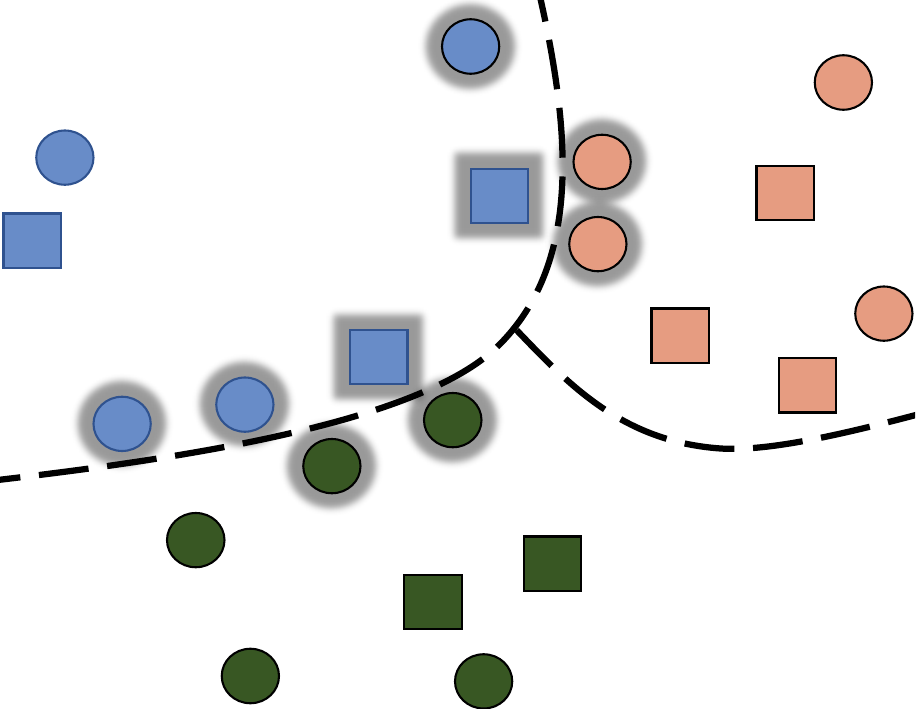}
        \caption{Visual modality representation spaces post-selective visual unlearning: cross-class samples are pushed to decision boundaries, risking performance degradation.}
        \label{chutian2}
    \end{subfigure}
    \begin{subfigure}[t]{0.49\linewidth}
        \centering
        \includegraphics[width=0.9\linewidth]{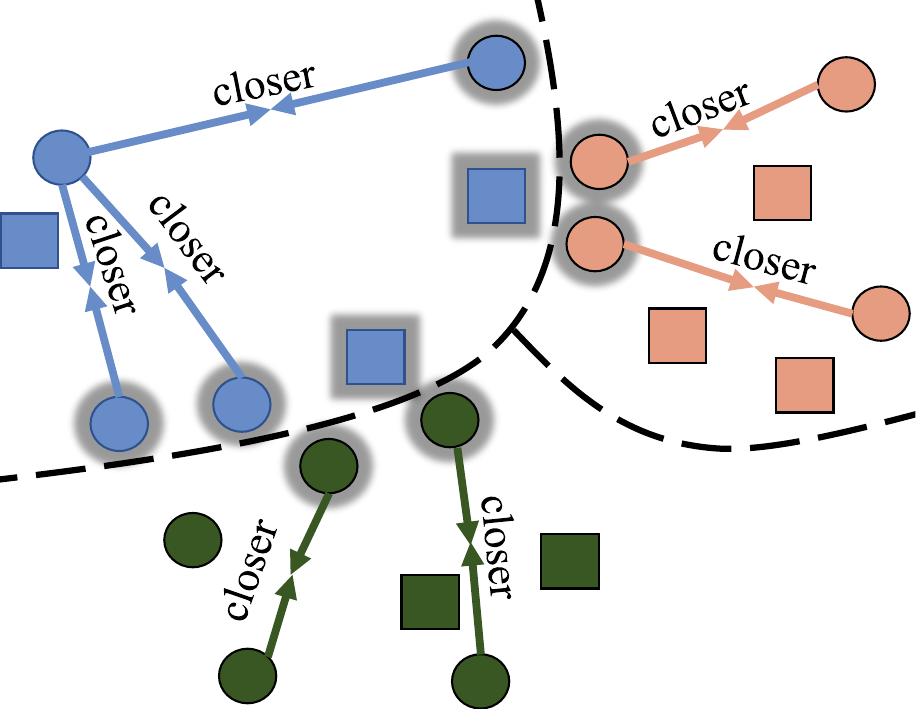}
        \caption{Visual modality representation spaces before dual-set contrastive separation, aligning current batch samples with same-class retain samples.}
        \label{chutian3}
    \end{subfigure}%
    \hfill
    \begin{subfigure}[t]{0.49\linewidth}
        \centering
        \includegraphics[width=0.9\linewidth]{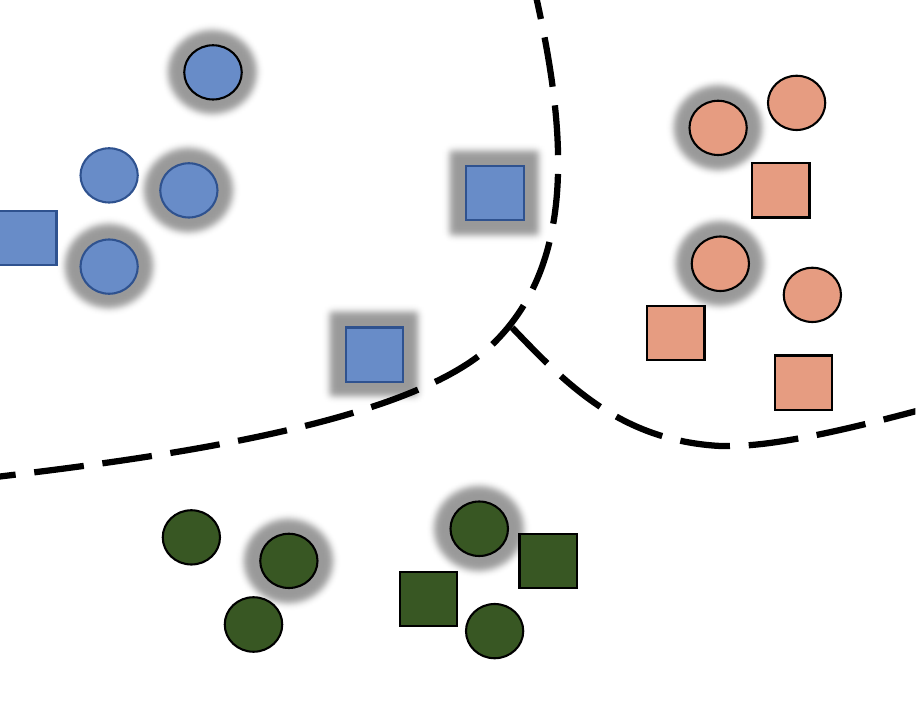}
        \caption{Visual modality representation spaces after dual-set contrastive separation, which tightens intra-class representations to preserve model performance.}
        \label{chutian4}
    \end{subfigure}
    \hfill
    \begin{subfigure}{1\linewidth}
        \centering
        \includegraphics[width=0.9\linewidth]{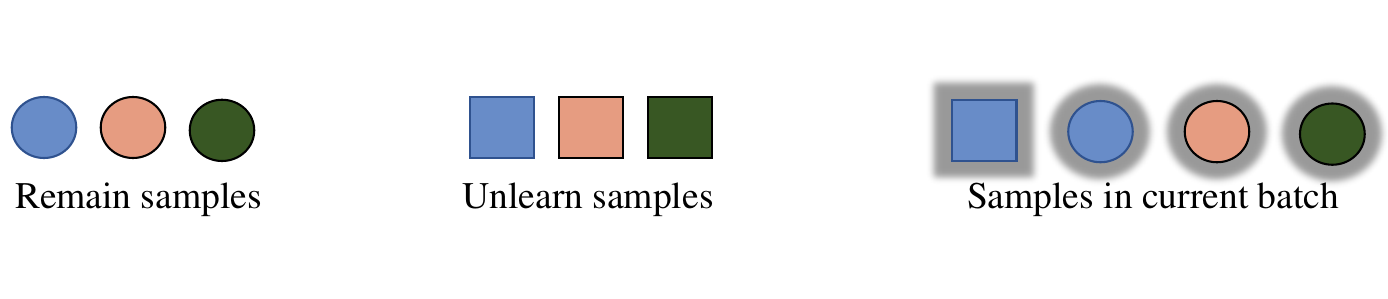}
        \label{fig:enter-label}
    \end{subfigure}
    \caption{Changes in the representation space after the application of selective visual unlearning and dual-set contrastive separation. Circles and squares represent the embeddings of remain samples and unlearning samples respectively. Different colors indicate distinct classes, and dotted lines illustrate the decision boundaries. We assume the model has been trained such that the embeddings of training samples are clustered according to their respective classes, as demonstrated in \cite{das2019separability}.}
    \label{fig:overall}
\end{figure}

During the unlearning process, we utilize the unlearn set as the anchor and the data in the retain set as positives and negatives. This approach for visual unlearning has an impact on the internal structure of each class in the retain set, thereby affecting the performance of the model on the retain set. To address the structural disruption of class representation spaces in retain set during visual unlearning, we propose dual-set contrastive separation (DCS). This module applies intra-modality contrastive learning on $D_r$ in the final $n$ unlearning iterations. We additionally introduce contrastive loss for both vision and audio within the retain set. For \({V}_{m}\), the loss function is:
\begin{equation}
\begin{split}
    \mathcal{L}_{DCS} = & \underbrace{-\frac{1}{{N}}\sum_{i = 1}^{{N}}\log\frac{\sum_{j\in{P}_i}\exp\left({m}_i\cdot {m}_{j^+}/\tau_{{DCS}}\right)}{\sum_{j\neq i}\exp\left({m}_i\cdot {m}_{j^-}/\tau_{{DCS}}\right)}}_{\mathcal{L}_v} \\
    & + \underbrace{-\frac{1}{{N}}\sum_{i = 1}^{{N}}\log\frac{\sum_{j\in{P}_i}\exp\left({n}_i\cdot {n}_{j^+}/\tau_{{DCS}}\right)}{\sum_{j\neq i}\exp\left({n}_i\cdot{n}_{j^-}/\tau_{{DCS}}\right)}}_{\mathcal{L}_a}
\end{split}
\label{con:DCS_loss_annotated}
\end{equation}

Fig \ref{chutian1} and Fig \ref{chutian2} illustrate that our selective visual unlearning module can dissociate visual representations of unlearn samples from their original semantics, but it will also affect retain samples. Fig \ref{chutian3} and Fig \ref{chutian4} illustrate that our dual-set contrastive separation module mitigates structural disturbances between the unlearn set and the retain set, thereby preserving the stability of intra-class representations and ensuring robust model performance.

\subsection{Optimization}
The proposed framework integrates multiple objectives through a weighted aggregate loss function, optimized using stochastic gradient descent:
\begin{equation}
    \mathcal{L} = \alpha\mathcal{L}_{SVU} + \beta\mathcal{L}_{CKR} + \gamma\mathcal{L}_{DCS}
\end{equation}
Here $\mathcal{L}_{SVU}$ enables selective visual unlearning by disrupting the representation spaces of unlearned visual data while safeguarding audio modality performance. $\mathcal{L}_{CKR}$ ensures cross-modal knowledge by aligning visual and audio embeddings through contrastive learning. $\mathcal{L}_{DCS}$ sustains the model's discriminative ability on retained data by optimizing intra-modality representations. Our combined loss function effectively balances visual unlearning with performance preservation, eliminating targeted visual data influence while maintaining general functionality and cross-modal knowledge. This balanced framework is critical for practical machine unlearning, ensuring complete data removal without sacrificing model utility.

\section{Experiment}
\begin{table*}[ht!]
  \centering
  \small
  \setlength{\aboverulesep}{0pt}
  \setlength{\belowrulesep}{0pt}
  \setlength{\arrayrulewidth}{0.8pt}
  \begin{tabular}{@{}c|c|c|c|c|c|c|c|c|cc|@{}} 
    \toprule
    FR& Set & Metric & Retrain & \textbf{CCU} & Finetune & NegGrad+ & Bad-T & SCRUB & SSD \\
    \midrule
    \multirow{11}{*}{5\%} & \multirow{5}{*}{Test} & $Acc_{all}$ & 0.7074 & \textbf{0.7251} & 0.7219 & 0.6397 & 0.6092 & 0.6430 & 0.5287 \\ 
    & & $Acc_{audio}$ & 0.1579 & \textbf{0.3087} & 0.2691 & 0.1599 & 0.0913 & 0.1874 & 0.1341 \\ 
    & & $Acc_{visual}$ & 0.5015 & \textbf{0.6219} & 0.6193 & 0.5842 & 0.5949 & 0.4653 & 0.5231 \\ 
    \cline{3-10}
    & & \multicolumn{1}{c|}{\cellcolor{gray!20}UPG(\%)$\uparrow$} & \multicolumn{1}{c|}{\cellcolor{gray!20}0.0} & \multicolumn{1}{c|}{\cellcolor{gray!20}\textbf{+2.5}} & \multicolumn{1}{c|}{\cellcolor{gray!20}+2.0} & \multicolumn{1}{c|}{\cellcolor{gray!20}-9.6} & \multicolumn{1}{c|}{\cellcolor{gray!20}-13.9} & \multicolumn{1}{c|}{\cellcolor{gray!20}-9.1} & \multicolumn{1}{c}{\cellcolor{gray!20}-25.3} \\ 
    \cline{3-10}
    & & \multicolumn{1}{c|}{\cellcolor{gray!20}CMSG(\%)$\uparrow$} & \multicolumn{1}{c|}{\cellcolor{gray!20}0.0} & \multicolumn{1}{c|}{\cellcolor{gray!20}\textbf{+95.5}} & \multicolumn{1}{c|}{\cellcolor{gray!20}+70.4} & \multicolumn{1}{c|}{\cellcolor{gray!20}+1.3} & \multicolumn{1}{c|}{\cellcolor{gray!20}-42.2} & \multicolumn{1}{c|}{\cellcolor{gray!20}+18.7} & \multicolumn{1}{c}{\cellcolor{gray!20}-15.1} \\
    \cline{2 - 10} 
    & \multirow{3}{*}{Retain} & $Acc_{all}$ & 0.7648 & 0.7256 & \textbf{0.7531} & 0.6340 & 0.6211 & 0.6716 & 0.5191 \\ 
    & & $Acc_{audio}$ & 0.1794 & \textbf{0.3248} & 0.3206 & 0.1786 & 0.0983 & 0.2152 & 0.1345 \\ 
    & & $Acc_{visual}$ & 0.4840 & 0.6090 & \textbf{0.6491} & 0.5712 & 0.6007 & 0.4589 & 0.5166 \\ 
    \cline{2 - 10} 
    & \multirow{3}{*}{Unlearn} & $Acc_{all}$ & 0.7381 & 0.7217 & 0.7323 & 0.6567 & 0.6019 & 0.6739 & 0.5287 \\ 
    & & $Acc_{audio}$ & 0.1773 & 0.3343 & 0.2713 & 0.1627 & 0.0984 & 0.2134 & 0.1269 \\ 
    & & $Acc_{visual}$ & 0.4710 & 0.5996 & 0.6413 & 0.5789 & 0.5916 & 0.4595 & 0.5227 \\  
    \midrule
    \multirow{11}{*}{15\%} & \multirow{5}{*}{Test} & $Acc_{all}$ & 0.7104 & \textbf{0.7115} & 0.7110 & 0.6574 & 0.6190 & 0.6317 & 0.3379 \\ 
    & & $Acc_{audio}$ & 0.1361 & \textbf{0.2944} & 0.2493 & 0.1658 & 0.0982 & 0.2781 & 0.1093 \\ 
    & & $Acc_{visual}$ & 0.5961 & \textbf{0.6108} & 0.6017 & 0.6064 & 0.5958 & 0.5422 & 0.4108 \\ 
    \cline{3-10}
    & & \multicolumn{1}{c|}{\cellcolor{gray!20}UPG(\%)$\uparrow$} & \multicolumn{1}{c|}{\cellcolor{gray!20}0.0} & \multicolumn{1}{c|}{\cellcolor{gray!20}\textbf{+0.2}} & \multicolumn{1}{c|}{\cellcolor{gray!20}+0.1} & \multicolumn{1}{c|}{\cellcolor{gray!20}-7.4} & \multicolumn{1}{c|}{\cellcolor{gray!20}-12.9} & \multicolumn{1}{c|}{\cellcolor{gray!20}-11.0} & \multicolumn{1}{c}{\cellcolor{gray!20}-52.4} \\ 
    \cline{3-10}
    & & \multicolumn{1}{c|}{\cellcolor{gray!20}CMSG(\%)$\uparrow$} & \multicolumn{1}{c|}{\cellcolor{gray!20}0.0} & \multicolumn{1}{c|}{\cellcolor{gray!20}\textbf{+116.3}} & \multicolumn{1}{c|}{\cellcolor{gray!20}+83.2} & \multicolumn{1}{c|}{\cellcolor{gray!20}+21.8} & \multicolumn{1}{c|}{\cellcolor{gray!20}-27.8} & \multicolumn{1}{c|}{\cellcolor{gray!20}+104.3} & \multicolumn{1}{c}{\cellcolor{gray!20}-19.7} \\
    \cline{2 - 10} 
    & \multirow{3}{*}{Retain} & $Acc_{all}$ & 0.7656 & \textbf{0.7250} & 0.7257 & 0.6625 & 0.6215 & 0.6663 & 0.3278 \\ 
    & & $Acc_{audio}$ & 0.1724 & \textbf{0.3199} & 0.2747 & 0.1827 & 0.0984 & 0.2960 & 0.1122 \\ 
    & & $Acc_{visual}$ & 0.5912 & \textbf{0.6139} & 0.6709 & 0.5933 & 0.5574 & 0.5574 & 0.3995 \\ 
    \cline{2 - 10} 
    & \multirow{3}{*}{Unlearn} & $Acc_{all}$ & 0.7511 & 0.7296 & 0.7637 & 0.6617 & 0.6294 & 0.6518 & 0.3236 \\ 
    & & $Acc_{audio}$ & 0.1536 & 0.3240 & 0.2894 & 0.1715 & 0.0988 & 0.3057 & 0.1045 \\ 
    & & $Acc_{visual}$ & 0.5714 & 0.6074 & 0.6612 & 0.6013 & 0.6035 & 0.5505 & 0.3973 \\  
    \bottomrule
  \end{tabular}
  \caption{Performance Evaluation on AV-MNIST. ``FR" means the percentage of the unlearn set in the training set.}
  \label{tab:1}
\end{table*}

\begin{table*}[ht!]
  \centering
  \small
  \setlength{\aboverulesep}{0pt}
  \setlength{\belowrulesep}{0pt}
  \begin{tabular}{@{}c|c|c|c|c|c|c|c|c|cc|@{}}
    \toprule
    FR& Set & Metric & Retrain & \textbf{CCU} & Finetune & NegGrad+ & Bad-T & SCRUB & SSD \\
     \midrule
    \multirow{11}{*}{5\%} & \multirow{5}{*}{Test} & $Acc_{all}$ & 0.5816 & \textbf{0.5976} & 0.5859 & 0.5828 & 0.5726 & 0.5424 & 0.4317 \\ 
    & & $Acc_{audio}$ & 0.2359 & \textbf{0.2496} & 0.2174 & 0.2382 & 0.2149 & 0.1225 & 0.2310 \\ 
    & & $Acc_{visual}$ & 0.6082 & 0.6039 & 0.5669 & \textbf{0.6205} & 0.5933 & 0.5489 & 0.4208 \\ 
    \cline{3-10}
    & & \multicolumn{1}{c|}{\cellcolor{gray!20}UPG(\%)$\uparrow$} & \multicolumn{1}{c|}{\cellcolor{gray!20}0.0} & \multicolumn{1}{c|}{\cellcolor{gray!20}\textbf{+2.8}} & \multicolumn{1}{c|}{\cellcolor{gray!20}+0.7} & \multicolumn{1}{c|}{\cellcolor{gray!20}+0.2} & \multicolumn{1}{c|}{\cellcolor{gray!20}-1.5} & \multicolumn{1}{c|}{\cellcolor{gray!20}-6.7} & \multicolumn{1}{c}{\cellcolor{gray!20}-25.8} \\ 
    \cline{3-10}
    & & \multicolumn{1}{c|}{\cellcolor{gray!20}CMSG(\%)$\uparrow$} & \multicolumn{1}{c|}{\cellcolor{gray!20}0.0} & \multicolumn{1}{c|}{\cellcolor{gray!20}\textbf{+5.8}} & \multicolumn{1}{c|}{\cellcolor{gray!20}-7.8} & \multicolumn{1}{c|}{\cellcolor{gray!20}+1.0} & \multicolumn{1}{c|}{\cellcolor{gray!20}-8.9} & \multicolumn{1}{c|}{\cellcolor{gray!20}-48.1} & \multicolumn{1}{c}{\cellcolor{gray!20}-2.1} \\
    \cline{2 - 10}
    & \multirow{3}{*}{Retain} & $Acc_{all}$ & 0.6314 & \textbf{0.6592} & 0.6189 & 0.6381 & 0.6295 & 0.6427 & 0.4598 \\ 
    & & $Acc_{audio}$ & 0.1787 & \textbf{0.2378} & 0.1996 & 0.2163 & 0.1850 & 0.1185 & 0.2198 \\ 
    & & $Acc_{visual}$ & 0.6555 & 0.6572 & 0.6138 & \textbf{0.6593} & 0.6046 & 0.6471 & 0.4703 \\ 
    \cline{2 - 10}
    & \multirow{3}{*}{Unlearn} & $Acc_{all}$ & 0.6274 & 0.6015 & 0.5484 & 0.5882 & 0.5994 & 0.5710 & 0.3941 \\ 
    & & $Acc_{audio}$ & 0.1659 & 0.2265 & 0.1851 & 0.2136 & 0.1907 & 0.1210 & 0.2148 \\ 
    & & $Acc_{visual}$ & 0.6369 & 0.6031 & 0.5531 & 0.6203 & 0.5891 & 0.5480 & 0.3976 \\  
    \midrule
    \multirow{11}{*}{15\%} & \multirow{5}{*}{Test} & $Acc_{all}$ & 0.5707 & \textbf{0.6302} & 0.5773 & 0.5429 & 0.4223 & 0.5520 & 0.4833 \\ 
    & & $Acc_{audio}$ & 0.2246 & \textbf{0.2960} & 0.2174 & 0.2250 & 0.2065 & 0.0921 & 0.2039 \\ 
    & & $Acc_{visual}$ & 0.5961 & \textbf{0.6226} & 0.5486 & 0.5135 & 0.4411 & 0.5492 & 0.5135 \\ 
    \cline{3-10}
    & & \multicolumn{1}{c|}{\cellcolor{gray!20}UPG(\%)$\uparrow$} & \multicolumn{1}{c|}{\cellcolor{gray!20}0.0} & \multicolumn{1}{c|}{\cellcolor{gray!20}\textbf{+10.4}} & \multicolumn{1}{c|}{\cellcolor{gray!20}+1.2} & \multicolumn{1}{c|}{\cellcolor{gray!20}-4.9} & \multicolumn{1}{c|}{\cellcolor{gray!20}-26.0} & \multicolumn{1}{c|}{\cellcolor{gray!20}-3.3} & \multicolumn{1}{c}{\cellcolor{gray!20}-15.3} \\ 
    \cline{3-10}
    & & \multicolumn{1}{c|}{\cellcolor{gray!20}CMSG(\%)$\uparrow$} & \multicolumn{1}{c|}{\cellcolor{gray!20}0.0} & \multicolumn{1}{c|}{\cellcolor{gray!20}\textbf{+31.8}} & \multicolumn{1}{c|}{\cellcolor{gray!20}-3.2} & \multicolumn{1}{c|}{\cellcolor{gray!20}+0.2} & \multicolumn{1}{c|}{\cellcolor{gray!20}-8.1} & \multicolumn{1}{c|}{\cellcolor{gray!20}-59.0} & \multicolumn{1}{c}{\cellcolor{gray!20}-9.2} \\
    \cline{2 - 10}
    & \multirow{3}{*}{Retain} & $Acc_{all}$ & 0.6295 & \textbf{0.7145} & 0.5980 & 0.5617 & 0.4283 & 0.6591 & 0.5045 \\ 
    & & $Acc_{audio}$ & 0.1790 & \textbf{0.2960} & 0.2023 & 0.2078 & 0.1898 & 0.0789 & 0.1843 \\ 
    & & $Acc_{visual}$ & 0.6500 & \textbf{0.6979} & 0.5646 & 0.5316 & 0.4568 & 0.6589 & 0.5322 \\ 
    \cline{2 - 10}
    & \multirow{3}{*}{Unlearn} & $Acc_{all}$ & 0.6168 & 0.6518 & 0.5348 & 0.5057 & 0.3901 & 0.5924 & 0.4671 \\ 
    & & $Acc_{audio}$ & 0.1639 & 0.2807 & 0.1932 & 0.2026 & 0.1934 & 0.1002 & 0.1753 \\ 
    & & $Acc_{visual}$ & 0.6310 & 0.6127 & 0.5031 & 0.4750 & 0.4095 & 0.5460 & 0.4992 \\  
    \bottomrule
  \end{tabular}
  \caption{Performance Evaluation on AVE.}
  \label{tab:2}
\end{table*}

\subsection{Experiment Setup}
\subsubsection{Datasets}
\textbf{AV-MNIST} \cite{10.1007/978-3-030-11024-6_44} is a multimodal dataset with 70,000 samples, formed by pairing MNIST handwritten digit images \cite{726791} with corresponding TIDIGITS spoken digit audio \cite{leonard1993tidigits}. \textbf{AVE} \cite{Tian_2018_ECCV} is an audio-visual event localization dataset containing 4,413 ten-second video clips, annotated with 28 event classes and frame-level labels. Sourced from YouTube, it follows the splits and preprocessing in \cite{Tian_2018_ECCV}. \textbf{UPMC-Food 101} \cite{bossard14} is a large-scale food classification dataset with ~100,000 samples across 101 categories; a 25-category subset is used here. Each instance includes an image and textual descriptions, with performance evaluated via classification accuracy following the preprocessing in \cite{guo2024classifier}.

\subsubsection{Experimental Settings}
For the \textbf{AV-MNIST} and \textbf{AVE} datasets, we adopt identical unlearn set configurations, comprising 5\% or 15\% of randomly selected training samples, with 95\% or 85\% retained as the retain set. For \textbf{AV-MNIST}, we follow \cite{li2023boosting}, using ResNet18-based audio and visual encoders, employing joint training and early fusion. For \textbf{AVE}, we adhere to \cite{Tian_2018_ECCV}, applying the same dataset division strategy. For the \textbf{UPMC-Food 101} dataset, we use encoders and classifiers from \cite{guo2024classifier}, with unlearn sets of 15\% or 30\% and retain sets of 85\% or 70\%; detailed results are in the supplementary material. Experiments with state-of-the-art unlearning methods validate our framework’s efficacy in addressing visual unlearning challenges in multimodal scenarios.

\subsubsection{Baseline Unlearning Methods}
We compare our unlearning framework with five approximate unlearning methods (abbreviated in bold), originally designed for unlearning in standard single-modal models:
(1) \textbf{Finetune} \cite{warnecke2021machine} leverages catastrophic forgetting to retrain the original model on retain samples. (2) \textbf{NegGrad+}: An enhanced version of NegGrad \cite{9157084} (which performs gradient ascent on unlearn samples) with added Cross-Entropy loss for stability. (3) \textbf{Bad-T} \cite{Chundawat_Tarun_Mandal_Kankanhalli_2023} uses knowledge distillation \cite{hinton2015distilling} to unlearn specific samples.
(4) \textbf{SCRUB} \cite{10.5555/3666122.3666217} adopts a student-teacher framework for unlearning. (5) \textbf{SSD} \cite{10.1609/aaai.v38i11.29092} prunes model parameters associated with unlearn samples. For fairness, all methods include an early-stopping condition: training stops when the model’s accuracy on the visual modality of the unlearn set falls below that on the visual modality of the test set.

\subsubsection{Evaluation Metrics}
\textbf{Model Performance} is evaluated using the accuracy of the unlearned model on the test set ${D}_{\text{test}}$ and the retain set ${D}_r$. The accuracy of the unlearned model on both datasets must not be greatly lower than that of the retrained model. To facilitate a more intuitive comparison, we introduce the \textbf{UPG (Utility Performance Gain)}:
$$ \text{UPG}_{\text{method}} = \left( \frac{\text{Acc}_{all,\text{ method}}^{\text{Test}}}{\text{Acc}_{all\text{, Retrain}}^{\text{Test}}} - 1 \right) \times 100\% $$
To assess \textbf{Cross-modal Knowledge Retention}, we also evaluate the accuracy of the unlearned model on other modalities. Higher accuracy indicates minimal disruption to cross-modal knowledge. To make the results more intuitive, we introduce the \textbf{CMSG (Cross-Modal Stability Gain)}: 
$$ \text{CMSG}_{\text{method}} = \left( \frac{\text{Acc}_{audio, \text{ method}}^{\text{Test}}}{\text{Acc}_{audio \text{, Retrain}}^{\text{Test}}} - 1 \right) \times 100\% $$

\begin{table}[ht!]
  \centering
  \small
  \belowrulesep=0pt
  \aboverulesep=0pt
  \begin{tabular}{@{}c|c|c|c|c@{}} 
    \toprule
    FR& Set & Acc & AV-MNIST & AVE  \\
    \midrule
    \multirow{9}{*}{5\%} & \multirow{3}{*}{Test} & $Acc_{all}$ & 0.7286 & 0.5877 \\ 
    & & $Acc_{audio}$ & 0.2955 & 0.2385 \\ 
    & & $Acc_{visual}$ & 0.6398 & 0.6184  \\ 
    \cline{2 - 5} 
    & \multirow{3}{*}{Retain} & $Acc_{all}$ & 0.7647 & 0.6377 \\ 
    & & $Acc_{audio}$ & 0.3350 & 0.2180 \\ 
    & & $Acc_{visual}$ & 0.6636 & 0.6591 \\ 
    \cline{2 - 5} 
    & \multirow{3}{*}{Unlearn} & $Acc_{all}$ & 0.7731 & 0.6312  \\ 
    & & $Acc_{audio}$ & 0.3378 & 0.1992\\ 
    & & $Acc_{visual}$ & 0.6727 & 0.6695 \\  
    \midrule
    \multirow{9}{*}{15\%} & \multirow{3}{*}{Test} & $Acc_{all}$ & 0.7315 & 0.5826 \\ 
    & & $Acc_{audio}$ & 0.2971 & 0.2113 \\ 
    & & $Acc_{visual}$ & 0.6403 & 0.6106 \\ 
    \cline{2 - 5} 
    & \multirow{3}{*}{Retain} & $Acc_{all}$ & 0.7602 & 0.6415 \\ 
    & & $Acc_{audio}$ & 0.3291 & 0.2193 \\ 
    & & $Acc_{visual}$ & 0.6642 & 0.6624 \\ 
    \cline{2 - 5} 
    & \multirow{3}{*}{Unlearn} & $Acc_{all}$ & 0.7628 & 0.6166\\ 
    & & $Acc_{audio}$ & 0.3346 & 0.2226 \\ 
    & & $Acc_{visual}$ & 0.6770 & 0.6434 \\  
    \bottomrule
  \end{tabular}
  \caption{Performance Evaluation Before Unlearning.}
  \label{tab:beforeul}
\end{table}

\textbf{Unlearning Efficiency} is measured by the time required to complete the unlearning process. Shorter execution times indicate higher computational efficiency.
In order to verify the effectiveness of unlearning, we carry out a \textbf{Membership Inference Attack (MIA)} \cite{7958568}. We suppose that an adversary has full access to the unlearned model and the training data. This is equivalent to simulating an administrator who employs the MIA to assess the efficacy of unlearning \cite{thudi2022necessity}. We present the \textit{Member Prediction Rate}, which is calculated as the ratio of the number of positive (member) predictions made by the MIA to the total number of tests. Specifically, for unlearn samples (regarded as non-members), this metric functions as the false positive rate (FPR), and for member samples, it serves as the true positive rate (TPR). For an unlearning algorithm to be effective, it is expected to result in a low member prediction rate for unlearn samples. A lower rate implies that the model is less prone to misclassifying them as members. We use the MIA framework from \cite{7958568} to compare our framework and other unlearning methods.

\subsection{Experiment Results}
\textbf{Model Performance} Table \ref{tab:beforeul} shows the model performance before unlearning. Table \ref{tab:1} presents the accuracy on the unlearn set, retain set, and test set for the AV-MNIST dataset. The maximum accuracy values in the test set and retain set are highlighted in bold. A successful unlearning method should achieve high test accuracy in both multi-modal and single-modal settings, while the visual modality accuracy on the unlearn set should be lower than that on the test set. \textbf{CCU} outperforms all other methods, even surpassing the Retraining method. In contrast, the Finetune method fails to unlearn effectively on AV-MNIST, as evidenced by its significantly higher visual accuracy on the unlearn set compared to the test set. When forget ratio reaches 15\%, our method even outperforms the model before unlearning, suggesting its applicability in more complex scenarios. The experiment results of UPMC-Food 101 can be seen in supplementary material.

\begin{figure}[ht!]
    \centering
    \begin{subfigure}[t]{0.49\linewidth}
        \centering
        \includegraphics[width=0.9\linewidth]{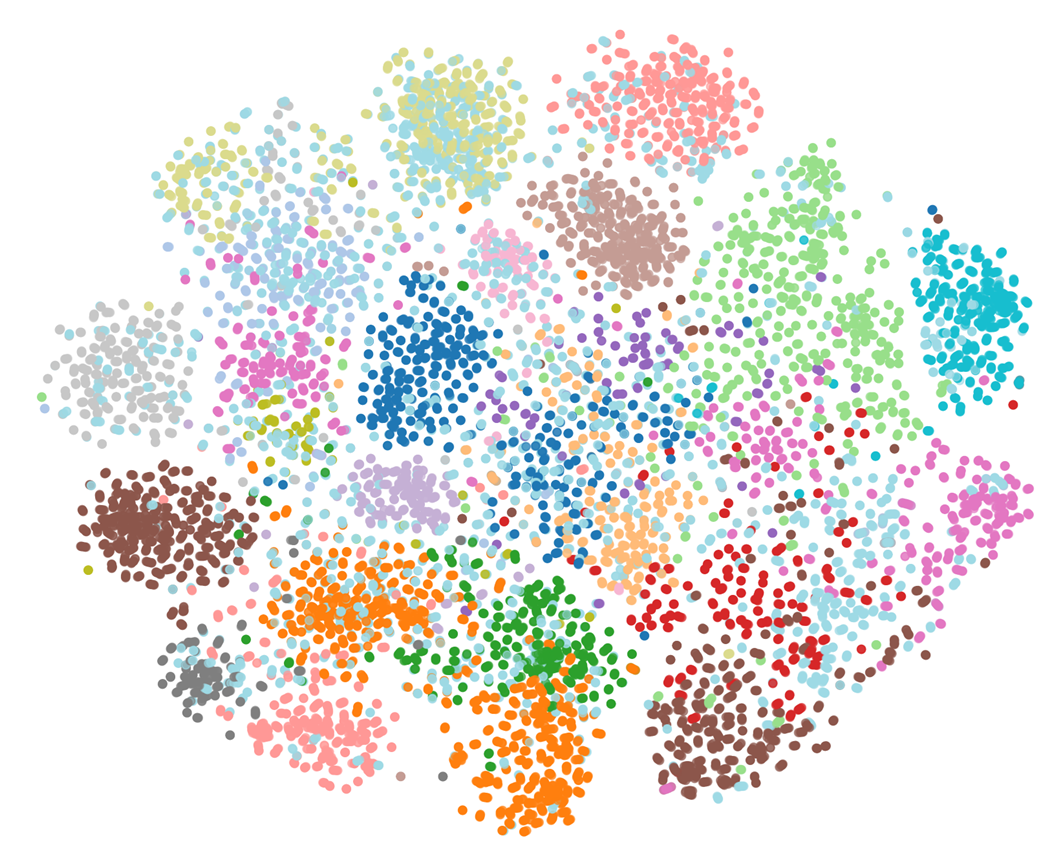}
        \caption{T-SNE visualization of pre-unlearning representation spaces, characterized by loose intra-class structure and indistinct inter-class boundaries.}
        \label{tsne1}
    \end{subfigure}%
    \hfill
    \begin{subfigure}[t]{0.49\linewidth}
        \centering
        \includegraphics[width=0.9\linewidth]{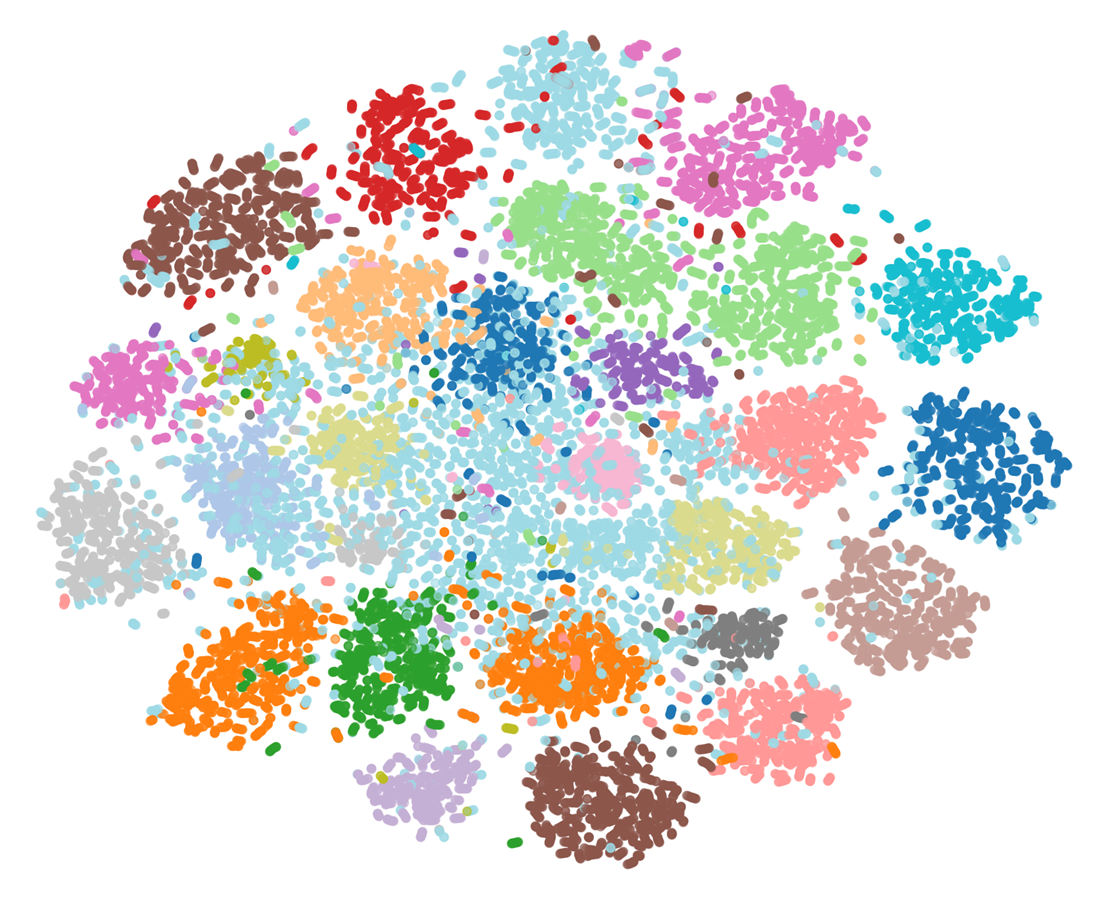}
        \caption{T-SNE visualization of representation spaces after CCU, showing tight intra-class structure and clear inter-class boundaries.}
        \label{tsne2}
    \end{subfigure}
    \caption{T-SNE visualization of representation spaces, where points of different colors denote representations of different classes.}
    \label{fig:TSNE}
\end{figure}

Table \ref{tab:2} shows performance on the AVE dataset. While NegGrad+ has higher visual accuracy on the test and retain sets for a 5\% forget ratio, our method is still the top performer overall. Additionally, our method achieves the highest audio accuracy, indicating its superior ability to preserve cross-modal knowledge, where other methods fall short. Fig \ref{tsne1} and Fig \ref{tsne2} show the T-SNE \cite{10.5555/2968618.2968725} visualization results of our method on the retain set. As can be seen in Fig \ref{tsne1}, the intra-class structure is loose, and the boundaries between classes are not obvious; in Fig \ref{tsne2}, the intra-class structure is tight, and the boundaries between classes are more obvious. These phenomena indicate that our method can achieve the goal of unlearning and preserving the model performance by modifying the representation spaces.

\textbf{Unlearning Efficiency.} Table \ref{tab:4} and Table \ref{tab:3} present the elapsed time for all unlearning methods on the AV-MNIST and AVE. The lowest values in the test set and remaining set are highlighted in bold. The elapsed time for the Retraining method is recorded when the retrained model achieves the highest accuracy on the test set. \textbf{CCU} demonstrates the fastest performance among all methods, indicating its efficiency in unlearning visual samples. This is because our method only uses a few samples from the remaining set for contrastive learning and modifies the representations of unlearning samples efficiently. This approach of using only a few samples from the remaining set for contrastive learning and efficiently modifying the representations of unlearning samples enables our method to achieve such high efficiency in the process of unlearning visual samples.
\begin{table}[ht!]
\setlength\tabcolsep{2pt}
    \centering
    \small
    \begin{tabular}{c|ccccccc}
        \toprule
         FR & Retrain & \textbf{CCU} & Finetune & NegGrad+ & Bad-T & SCRUB & SSD \\
        \midrule
        5\% & 3058.15 & \textbf{39.35} & 9537.15 & 67.49 & 105.29 & 9349.40 & 116.56\\
        15\% & 594.29 & \textbf{50.39} & 9750.52 & 63.67 & 127.78 & 10287.44 & 371.40\\
    \bottomrule
    \end{tabular}
  \caption{Processing time on AV-MNIST dataset (seconds).}
  \label{tab:4}
\end{table}

\begin{table}[ht!]
\setlength\tabcolsep{2pt}
    \centering
    \small
    \begin{tabular}{c|ccccccc}
        \toprule
         FR & Retrain & \textbf{CCU} & Finetune & NegGrad+ & Bad-T & SCRUB & SSD \\
        \midrule
        5\% & 780.42 & \textbf{7.35} & 14.26 & 99.01 & 8.64 & 153.30 & 26.50\\
        15\% & 513.92 & \textbf{8.04} & 14.01 & 16.35 & 35.68 & 291.70 & 13.31\\
        \bottomrule
    \end{tabular}
  \caption{Processing time on AVE dataset.}
  \label{tab:3}
\end{table}

\subsection{MIA Results}
Table \ref{tab:5} and Table \ref{tab:6} present the member prediction rates on the unlearning set for the MIA on the AV-MNIST and AVE datasets, respectively. An effective unlearning method should achieve a lower member prediction rate on the unlearning set. The lowest rates in the test set and remaining set are highlighted in bold. Our \textbf{CCU} framework achieves the lowest member prediction rates on both AV-MNIST and AVE, except for the case with a 15\% forget ratio on AVE. This demonstrates that our framework provides superior unlearning efficacy compared to other methods. This is because our method can maintain the representation space of the original model and only remove those samples that need to be unlearned.
\begin{table}[ht!]
\setlength\tabcolsep{2pt}
    \centering
    \small
    \begin{tabular}{c|ccccccc}
        \toprule
         FR & Retrain & \textbf{CCU} & Finetune & NegGrad+ & Bad-T & SCRUB & SSD \\
        \midrule
        5\% & 64.02 & \textbf{53.50} & 73.28 & 65.29 & 72.29 & 74.50 & 55.24\\
        15\% & 62.49 & \textbf{54.16} & 72.02 & 68.67 & 69.31 & 73.59 & 59.40\\
    \bottomrule
    \end{tabular}
    \caption{Member prediction rate on unlearn set (lower the better) of MIA on AV-MNIST dataset.}
    \label{tab:5}
\end{table}
\begin{table}[ht!]
\setlength\tabcolsep{2pt}
    \centering
    \small
    \begin{tabular}{c|ccccccc}
        \toprule
         FR & Retrain & \textbf{CCU} & Finetune & NegGrad+ & Bad-T & SCRUB & SSD \\
        \midrule
        5\% & 62.28 & 55.23 & 65.76 & 59.35 & 58.59 & 71.68 & \textbf{55.10}\\
        15\% & 65.20 & \textbf{53.41} & 63.53 & 59.62 & 62.70 & 72.83 & 57.95\\
        \bottomrule
    \end{tabular}
    \caption{Member prediction rate on unlearn set (lower the better) of MIA on AVE dataset.}
    \label{tab:6}
\end{table}

\section{Conclusion}
This paper proposes a novel cross-modal contrastive framework for visual unlearning in multimodal scenarios. The approach realizes visual forgetting via contrastive unlearning on visual data in the unlearn set, while preserving the astructure of the retain set. By strengthening the intra-class structure of retain samples, our \textbf{CCU} framework ensures model generalization; meanwhile, it effectively retains cross-modal shared knowledge through cross-modal knowledge retention. Comprehensive experiments show that \textbf{CCU} outperforms existing state-of-the-art unlearning algorithms in model performance, unlearning efficacy, and computational efficiency.

{
    \small
    \bibliographystyle{ieeenat_fullname}
    \bibliography{cvpr2026}
}

\end{document}